\documentclass[10pt,journal,compsoc]{IEEEtran}
\ifCLASSINFOpdf
\else
\fi

\usepackage{algorithmic}

\usepackage{booktabs}       
\usepackage{amsmath}
\usepackage{multirow}
\usepackage{lscape}
\usepackage{verbatim}
\usepackage{subcaption}
\usepackage[demo]{graphicx}
\usepackage{caption}
\graphicspath{ {./images/} }
\usepackage[sorting=none]{biblatex}
\begin{document}
%
\title{Modeling, Quantifying, and Predicting Subjectivity of Image Aesthetics}
%
%
%
%

\author {
    Hyeongnam Jang\textsuperscript{\rm 1},
    Yeejin Lee\textsuperscript{\rm 2},
    Jong-Seok Lee\textsuperscript{\rm 1}\\
    \textsuperscript{\rm 1} School of Integrated Technology, Yonsei University, Korea\\
    \textsuperscript{\rm 2} Department of Electrical and Information Engineering, \\Seoul National University of Science and Technology, Korea\\
    hyeongnam.jang@yonsei.ac.kr, yeejinlee@seoultech.ac.kr, jong-seok.lee@yonsei.ac.kr
}

%
%

\markboth{Journal of \LaTeX\ Class Files,~Vol.~14, No.~8, August~2015}%
{Shell \MakeLowercase{\textit{et al.}}: Bare Advanced Demo of IEEEtran.cls for IEEE Computer Society Journals}
%



\IEEEtitleabstractindextext{%
\begin{abstract}
Assessing image aesthetics is a challenging computer vision task. One reason is that aesthetic preference is highly subjective and may vary significantly among people for certain images. Thus, it is important to properly model and quantify such \textit{subjectivity}, but there has not been much effort to resolve this issue. In this paper, we propose a novel unified probabilistic framework that can model and quantify subjective aesthetic preference based on the subjective logic. In this framework, the rating distribution is modeled as a beta distribution, from which the probabilities of being definitely pleasing, being definitely unpleasing, and being uncertain can be obtained. We use the probability of being uncertain to define an intuitive metric of subjectivity. Furthermore, we present a method to learn deep neural networks for prediction of image aesthetics, which is shown to be effective in improving the performance of subjectivity prediction via experiments. We also present an application scenario where the framework is beneficial for aesthetics-based image recommendation.
\end{abstract}
}

\maketitle

\IEEEdisplaynontitleabstractindextext

%
\IEEEpeerreviewmaketitle

\ifCLASSOPTIONcompsoc
\IEEEraisesectionheading{\section{Introduction}\label{sec:introduction}}
\else
\section{Introduction}
\label{sec:introduction}
\fi

%
%
%
%
\IEEEPARstart{I}{mage} aesthetic assessment is to automatically evaluate the image in the aesthetic viewpoint, i.e., how aesthetically pleasing to human viewers an image will be. It is a challenging computer vision task since it requires to imitate high-level aesthetic perception of humans, but it can be useful in many applications including image search and retrieval, recommender systems, image enhancement, etc.

\begin{figure}[t]

\begin{subfigure}{\columnwidth}
  \centering
  \begin{subfigure}{0.39\columnwidth}
    \includegraphics[width=\linewidth]{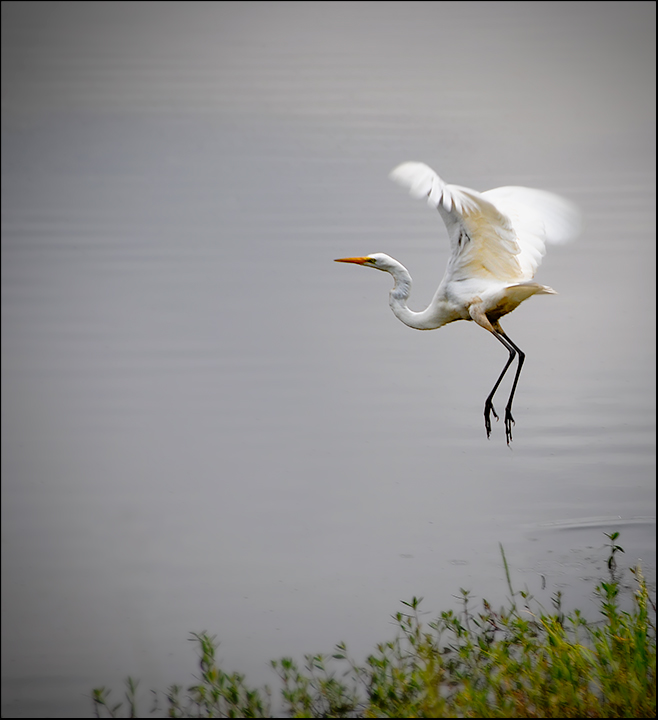}  
  \end{subfigure}
  \begin{subfigure}{0.60\columnwidth}
    \includegraphics[width=\linewidth]{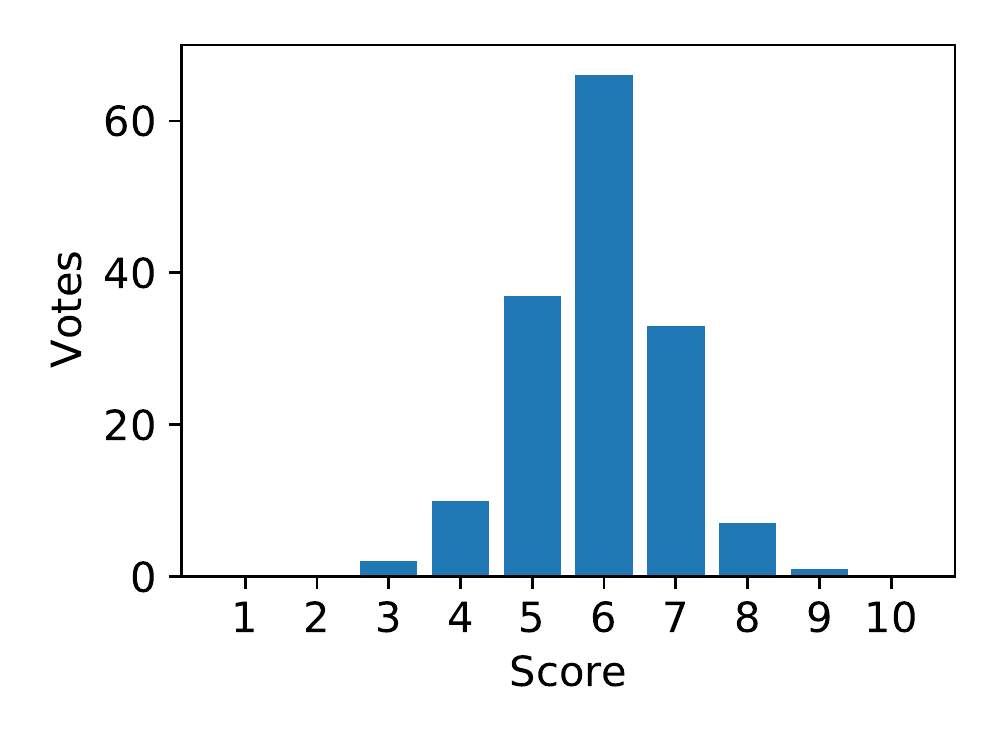}  
  \end{subfigure}
  \caption{Image having low subjectivity}
  \label{fig:votes1}
\end{subfigure}
\begin{subfigure}{\columnwidth}
  \centering
  \begin{subfigure}{0.39\columnwidth}
    \includegraphics[width=\linewidth]{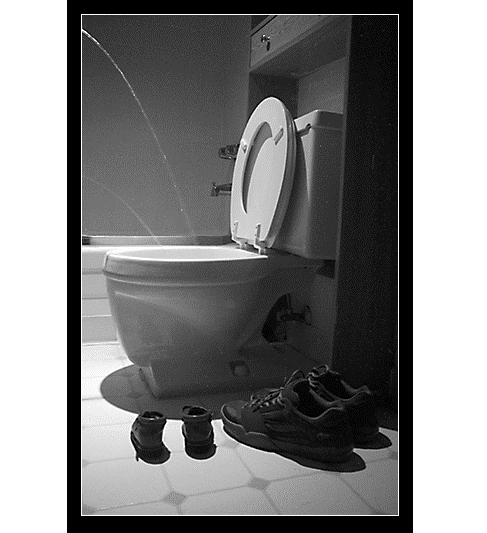}  
  \end{subfigure}
  \begin{subfigure}{0.60\columnwidth}
    \includegraphics[width=\linewidth]{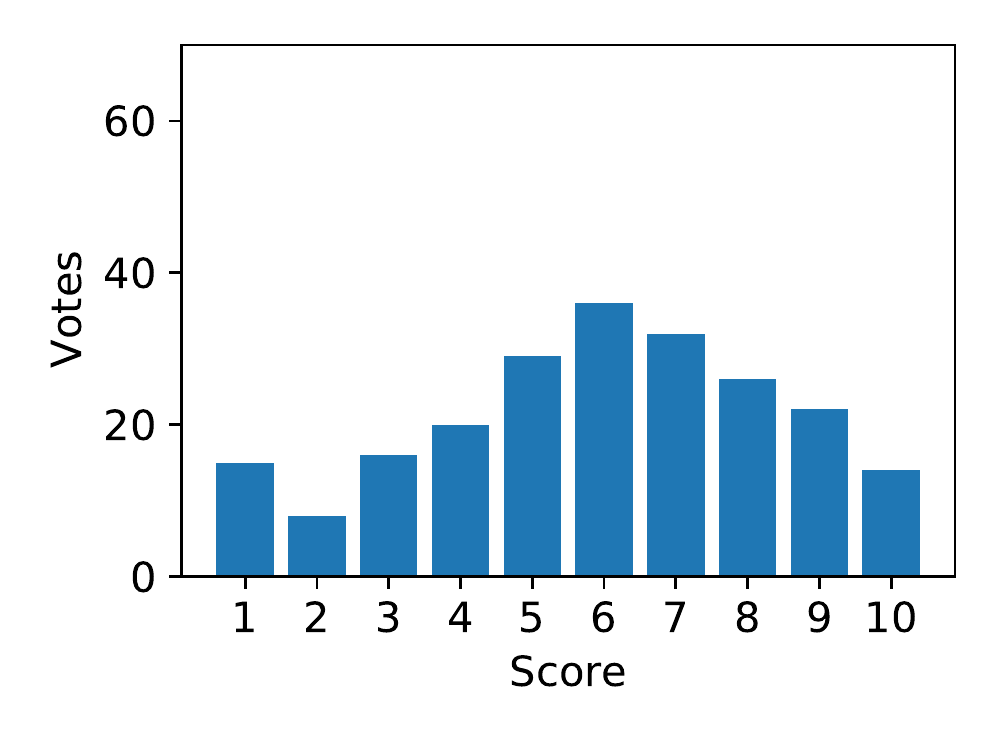}  
  \end{subfigure}
  \caption{Image having high subjectivity}
  \label{fig:votes2}
\end{subfigure}
\caption{Example images and their histograms of the rating scores given by raters. The two images have similar mean rating scores (about 5.92) but different degrees of subjectivity.}
\label{fig:lowhighsubjectivity}
\end{figure}

In order to obtain the aesthetic ground truth of an image, it is usual to ask a group of raters to provide aesthetic scores for the image. In the binary classification task, the image is considered as aesthetically pleasing if the mean score is higher than a threshold, and as unpleasing otherwise. It is also possible to set a regression task where the mean rating score is predicted. However, these tasks do not consider the diversity in the raters' opinions. In Fig. \ref{fig:lowhighsubjectivity}, two example images and their histograms of the rating scores from raters are shown, which are from the AVA dataset \cite{murray2012ava}. The two images have similar mean scores (about 5.92), so that they have the same target mean rating score for mean rating regression and the same class label for binary classification. However, these do not capture the different levels of diversity in the raters' opinions between the two cases. While the distribution of ratings in Fig. \ref{fig:votes1} is concentrated around the mean score, that in Fig. \ref{fig:votes2} is spread widely over the whole score range. Thus, in the case of Fig. \ref{fig:votes2}, the results of the mean rating regression and binary classification may be disagreed by a significant proportion of users.

Therefore, it is necessary to consider \emph{subjectivity} in aesthetic assessment. However, only a limited number of existing studies have dealt with the issue of subjectivity. In particular, the problem of quantifying the level of subjectivity still remains unanswered. A straightforward way is to compute the standard deviation (STD) (or variance) of the rating scores. However, it has an issue in terms of interpretability because it does not have an upper limit and the meaning of its scale is unclear. Similar metrics exist, such as mean absolute deviation around median \cite{kang2019predicting}, which suffer from the same problem.

In this paper, we propose a novel unified probabilistic framework for modeling and quantifying the subjectivity of image aesthetics based on the subjective logic \cite{josang2016subjective}. In this framework, the rating distribution of an image is modeled as a beta distribution, from which the probabilities of being definitely pleasing, being definitely unpleasing, and being uncertain can be obtained simultaneously. In particular, the probability of being uncertain defines an intuitive metric of subjectivity, named \emph{aesthetic uncertainty}. Unlike the existing subjectivity metrics, it is a probability measure, which can be easily interpreted. Furthermore, we present a method to train deep neural networks for effective prediction of the subjectivity metrics. We also present an application scenario where the framework is beneficial for aesthetics-based image recommendation.

\section{Related work}
\label{sec:relatedwork}

\subsection{Image aesthetic assessment}

In literature, three tasks have been mainly considered for automatic aesthetic assessment of images: binary classification, mean score regression, and rating distribution prediction. The binary classification task distinguishing pleasing vs. unpleasing images has been considered most popularly. There exist several methods, from those using handcrafted features \cite{li2009aesthetic, mavridaki2015comprehensive, sun2018photo} to deep learning-based methods \cite{lu2014rapid, Mai_2016_CVPR, Liu_2019_CVPR_Workshops, sheng2020revisiting}. To obtain more informative results than binary class information, the mean rating regression task has been addressed \cite{zeng2019unified, lee2019image, ke2021musiq}. Prediction of the whole score distribution has been also considered \cite{talebi2018nima, Chen_2020_CVPR, ching2020learning, she2021hierarchical}, which is the most challenging but has potential to provide the most comprehensive information regarding the aesthetic characteristics of the given image. In this paper, we consider all these tasks and also the subjectivity regression task.

\subsection{Subjectivity of image aesthetics}

Subjectivity is a clearly distinguished issue in image aesthetics compared to other image-based problems such as object classification. While the class of an object in an image can be objectively determined, subjective judgement is involved in image aesthetics, and thus an image preferred by certain viewers is not necessarily pleasing to some other viewers. Park et al. \cite{park2015consensus} modeled the human aesthetic evaluation process by a dynamic system and showed that the response time for aesthetic evaluation of an image is related to the subjectivity level of the image in terms of STD. Kim et al. \cite{kim2020objectivity} analyzed the relationship between subjectivity (expressed by STD) and user comments, which showed that several factors such as unusualness and coexistence of aesthetic merits and demerits are involved in determining the level of subjectivity of an image. There also exist studies suggesting personalized image aesthetic assessment techniques that reflect the information of a specific user \cite{ren2017personalized, wang2018collaborative, li2019personality, lee2019image, zhu2020personalized, lv2021user, yang2022personalized}.

The rating distribution itself can give the information regarding subjectivity (as in Fig. \ref{fig:lowhighsubjectivity}) but only implicitly \cite{kang2019predicting}. Therefore, there is a need to explicitly quantify subjectivity as a scalar value. However, there is not much progress in this research direction. STD has been mostly used \cite{talebi2018nima, jang2021analysis}. Kang et al. \cite{kang2019predicting} defined additional subjectivity metrics including the mean absolute deviation around the median (MAD), distance to uniform distribution (DUD), and distance from the maximum entropy distribution (MED). However, all these metrics have a limitation in interpretability because they have neither upper limits nor interpretable scales (except 0). Our work addresses this issue and proposes an intuitive metric of subjectivity.

\begin{figure}[t]
\centering
\begin{subfigure}{0.51\columnwidth}
  \centering
  \includegraphics[width=\linewidth]{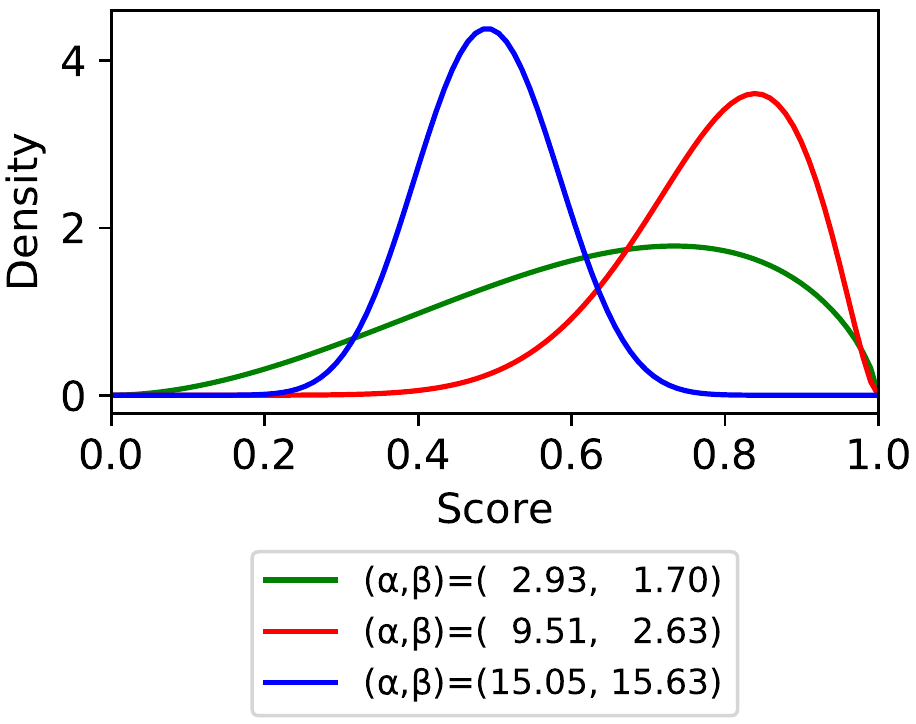}  
  \caption{Beta distribution}
  \label{fig:betapdfs}
\end{subfigure}
\begin{subfigure}{0.48\columnwidth}
  \centering
  \includegraphics[width=\linewidth]{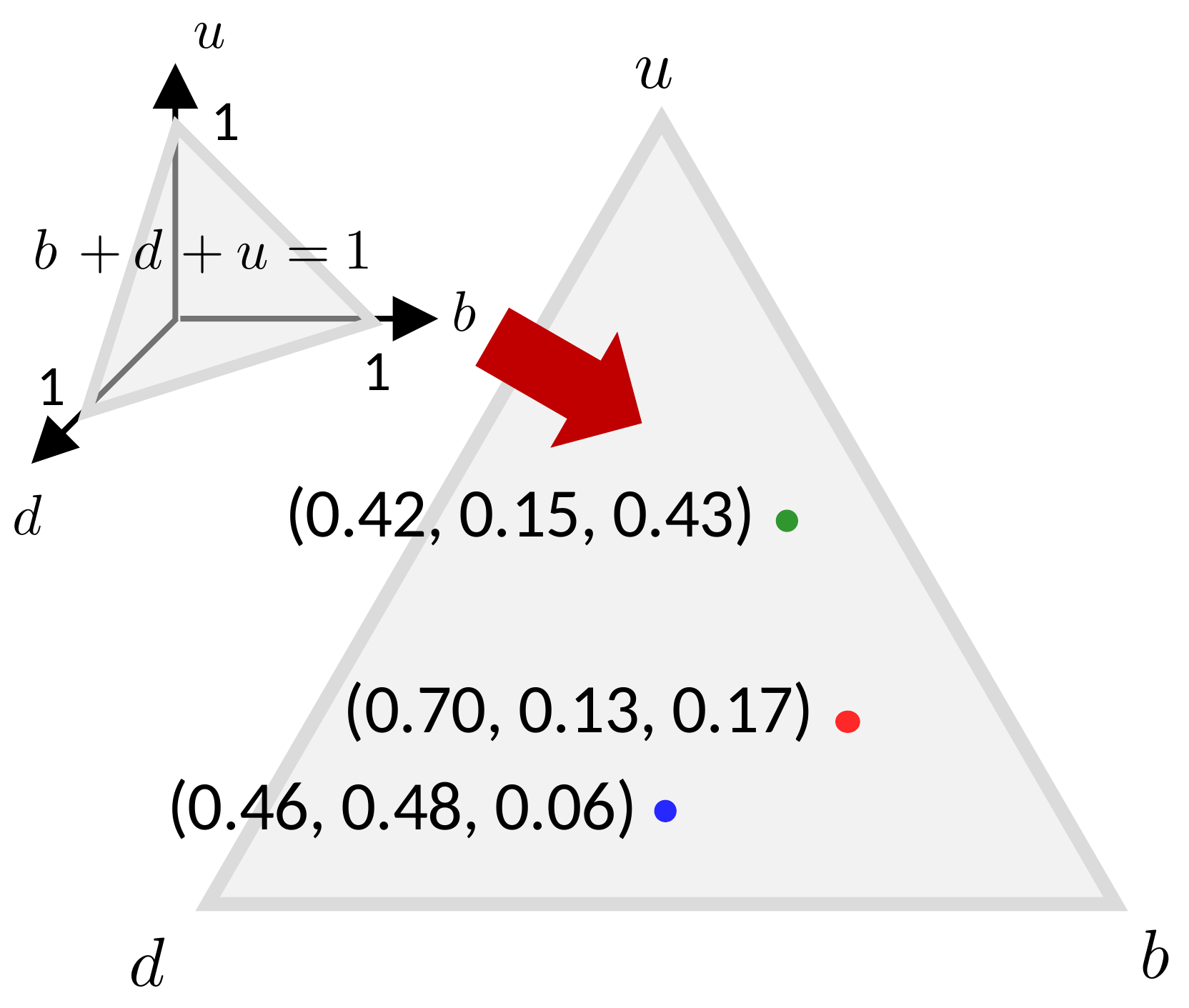}  
  \caption{Triple $(b,d,u)$}
  \label{fig:triangle}
\end{subfigure}
\caption{Image aesthetics via the subjectivity modeling. The fitted beta distributions and the corresponding triples of $(b,d,u)$ on the equilateral triangle are shown.}
\label{fig:triangles}
\end{figure}

\section{Proposed method for subjectivity modeling}
\label{sec:subjectivitymodeling}

The proposed unified framework is based on the subjective logic \cite{josang2016subjective}. The subjective logic is a probabilistic reasoning for modeling subjective opinions involving uncertainty. An aesthetic rating chosen from a range of scores (e.g., 1 to 10) is a multinomial opinion. However, if the rating scale is normalized between 0 and 1, the rating can be considered as a binomial opinion indicating the subjective belief about aesthetic pleasingness of the image. Then, this binomial opinion is represented by three components \cite{josang2016subjective}: $b$ (belief mass), $d$ (disbelief mass), and $u$ (uncertainty mass), which correspond to the probabilities of being definitely pleasing, being definitely unpleasing, and being uncertain, respectively. These components have values between 0 and 1, and satisfy $b+d+u=1$. As a result, the aesthetics of an image can be represented as a point on a equilateral triangle as shown in Fig. \ref{fig:triangle}.

The binomial opinion can be modeled by a beta distribution whose probability density function (PDF) is given by
\begin{equation}
f(x;\alpha,\beta)=\frac{1}{B(\alpha,\beta)}x^{\alpha-1}(1-x)^{\beta-1},
\label{eq:betadistribution}
\end{equation}
where 0 $\leq$ $x$ $\leq$ 1, $\alpha$ and $\beta$ are shape parameters, and $B(\alpha,\beta)$ is a normalization constant ensuring $\int_0^1 f(x;\alpha,\beta) dx = 1$. When the rating distribution of an image is given, a beta distribution is fitted to the distribution by finding the optimal values of $\alpha$ and $\beta$ that minimize the difference (e.g., earth movers' distance (EMD)) between the given and fitted distributions.

Finally, the three probabilities ($b$, $d$, and $u$) can be obtained from the fitted values of $\alpha$ and $\beta$ as follows \cite{josang2016subjective}:

\begin{equation}
\begin{aligned}
b= &\frac{\alpha -1}{\alpha +\beta },\\
d= &\frac{\beta -1}{\alpha +\beta },\\
u= &\frac{2}{\alpha +\beta }.
\end{aligned}
\label{eq:bdu2}
\end{equation}

Fig. \ref{fig:triangles} shows examples of fitted beta distributions and their representations in the equilateral triangle of $b$, $d$, and $u$. The image corresponding to the red-colored distribution in Fig. \ref{fig:betapdfs} would be considered as aesthetically pleasing by most people, which is reflected in the large value of $b$ and the small value of $u$ in Fig. \ref{fig:triangle}. The blue-colored case in Fig. \ref{fig:betapdfs} is judged to have an intermediate level of aesthetics without much disagreement among people, which is represented by the small value of $u$ and the similar values of $b$ and $d$ in Fig. \ref{fig:triangle}. The green-colored case would be classified as pleasing if binary classification is performed, but it involves high subjectivity; thus, $u$ appears to be large while $b>d$.

\begin{figure}[t]
\begin{subfigure}{0.49\columnwidth}
  \centering
  \includegraphics[width=\linewidth]{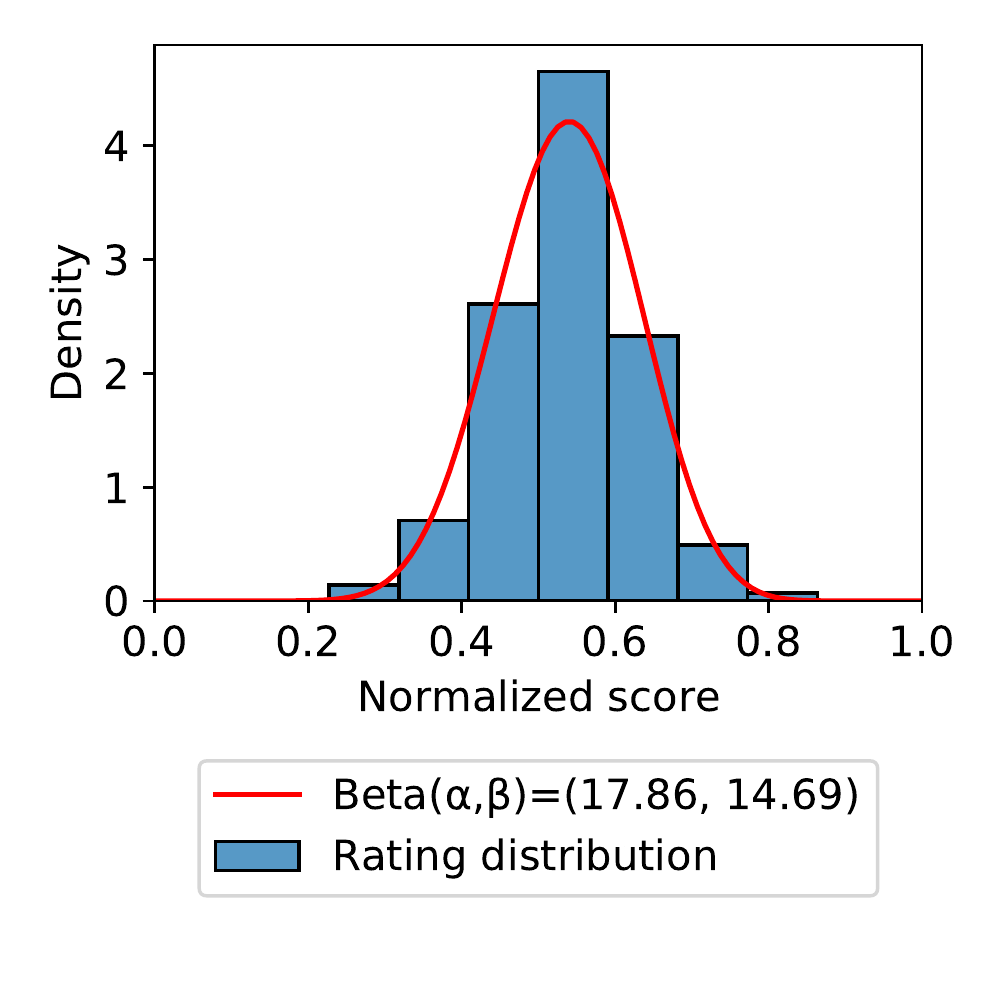}  
  \caption{Unimodal}
  \label{fig:monopole}
\end{subfigure}
\begin{subfigure}{0.49\columnwidth}
  \centering
  \includegraphics[width=\linewidth]{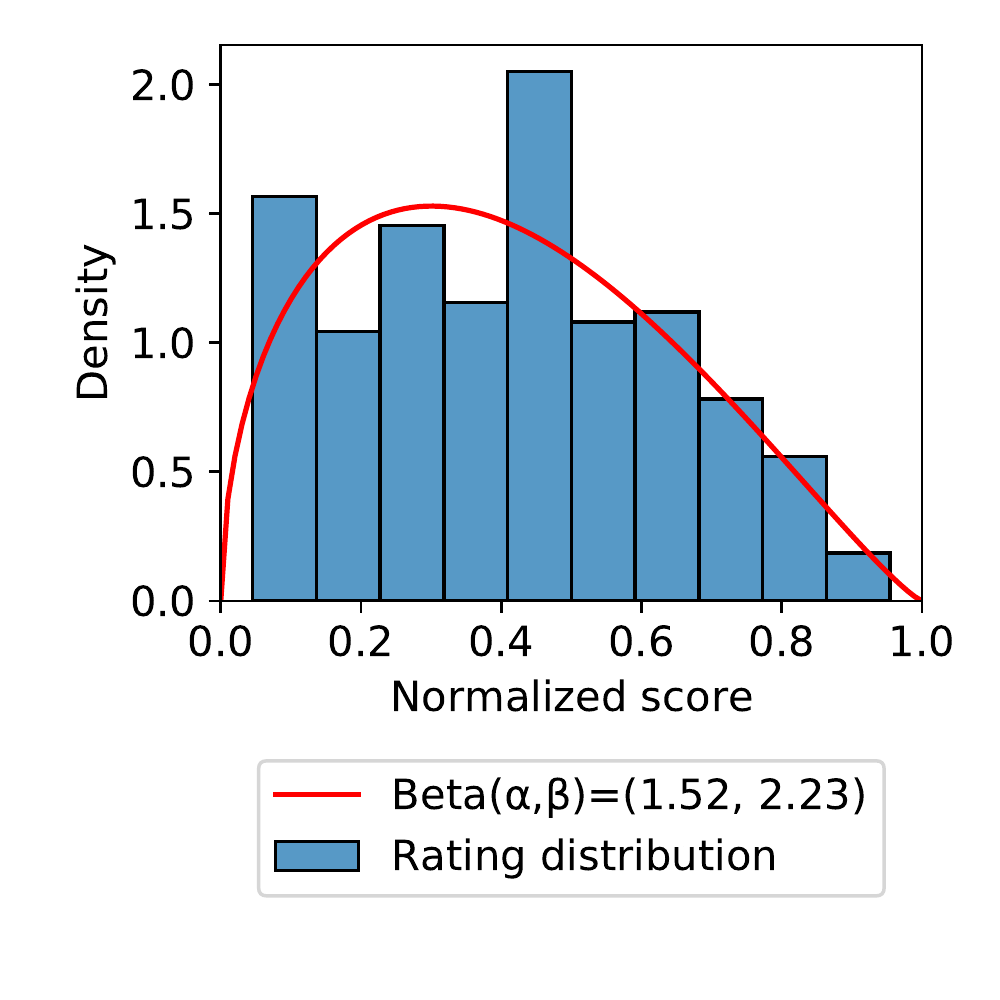}  
  \caption{Multimodal}
  \label{fig:multipole}
\end{subfigure}

\caption{Examples of unimodal and multimodal aesthetic rating distributions (after score normalization) and the fitted beta distributions.}
\label{fig:monomultipoles}
\end{figure}

The beta distribution is unimodal, while an original rating distribution may be multimodal. An example case is shown in Fig. \ref{fig:multipole}, along with the fitted beta distribution. For the images in the AVA dataset \cite{murray2012ava} used in our experiments, we conduct the dip test of unimodality \cite{hartigan1985dip} and find that 94.56\% of the images have unimodal distributions, whereas only 5.14\% and 0.30\% are bimodal and trimodal, respectively. Note that some of the multimodal distributions may be due to noise in the ratings, which can be reduced by the unimodal modeling. Thus, we can say that fitting to beta distributions is reasonable.

\subsection{Aesthetic uncertainty (AesU)}

We define a new metric quantifying the level of subjectivity, called aesthetic subjectivity (AesU), by $u$ obtained from the subjectivity modeling. It has several advantages compared to the existing metrics. Since AesU is a probability within [0, 1], one can intuitively grasp the level of subjectivity, whereas the scales of the existing metrics are not well defined and thus not sufficiently intuitive. In addition, AesU can be interpreted together with the other two probabilities ($b$ and $u$) as discussed in Fig. \ref{fig:triangle}, which is difficult with the mean and STD (or, MAD, DUD, MED) of ratings.

\begin{figure}[t]
\centering
\includegraphics[width=0.87\columnwidth]{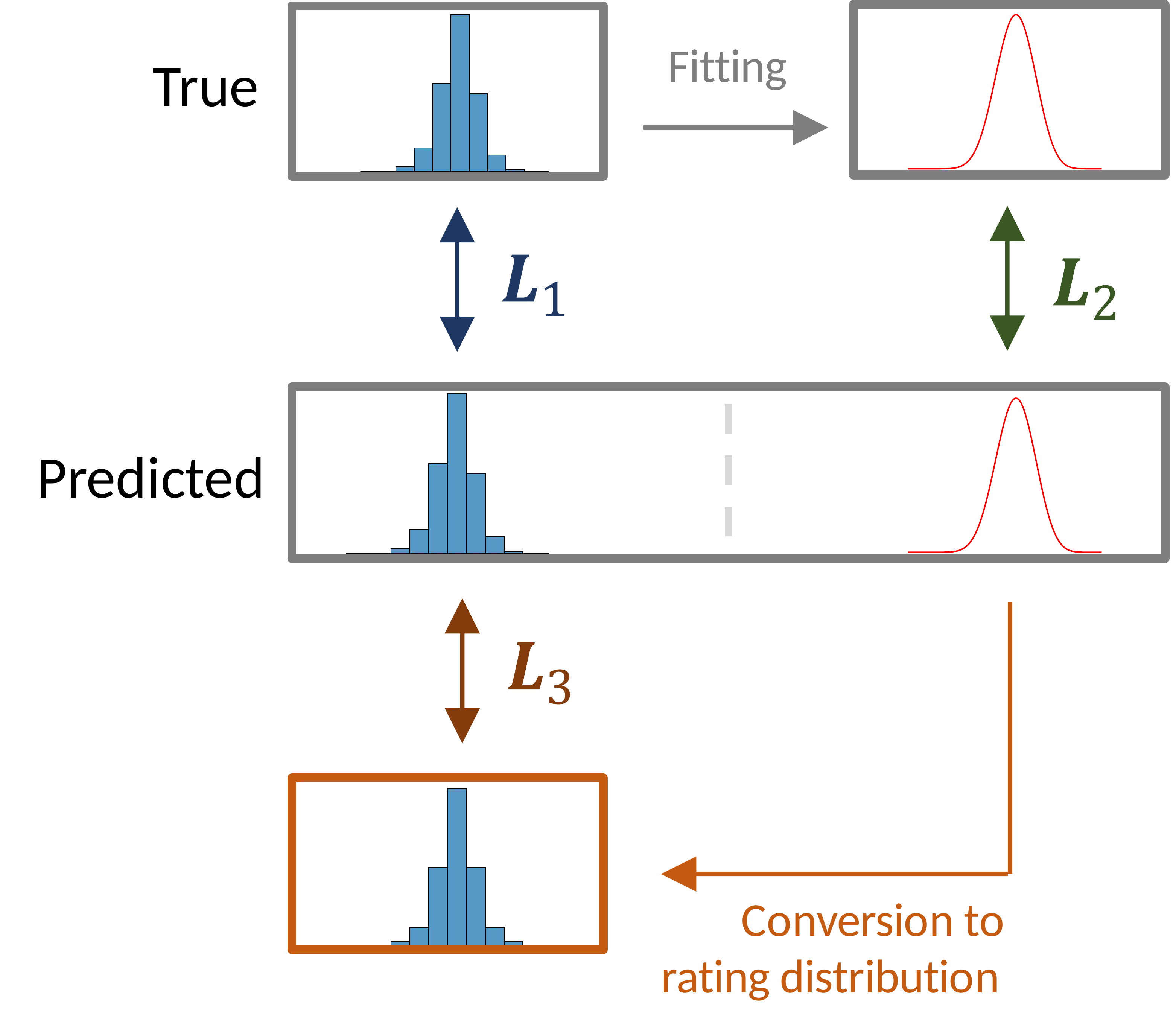}  
\caption{Schematic diagram of the losses used in the proposed simultaneous learning. The learned model predicts both the rating distribution and the shape parameters of the fitted beta distribution, from which $L_1$ and $L_2$ are computed, respectively. The predicted shape parameters are used for conversion to a rating distribution, for which $L_3$ is computed. The total loss is a weighted sum of $L_1$, $L_2$, and $L_3$.}
\label{fig:loss}
\end{figure}

\begin{table*}[t]
\centering
\begin{tabular}{@{}cc|c|ccccc@{}}
\toprule
Approach & Backbone & Performance measure & STD & MAD & MED & DUD & AesU \\ \midrule
\multirow{4}{*}{\begin{tabular}[c]{@{}c@{}}Conventional\\(rating distribution learning)\end{tabular}} & VGG16 & \multirow{8}{*}{PLCC ↑} & 0.2314 & 0.2041 & 0.3065 & 0.3188 & 0.2231 \\
 & ResNet50 &  & 0.2653 & 0.2539 & 0.3334 & 0.3627 & 0.2483 \\
 & ConvNeXT-T &  & 0.2930 & 0.2898 & 0.3643 & 0.3900 & 0.2750 \\
 & HLAGCN &  & 0.2399 & 0.1896 & 0.3347 & 0.3784 & 0.1638 \\ \cmidrule(r){1-2} \cmidrule(l){4-8} 
\multirow{4}{*}{\begin{tabular}[c]{@{}c@{}} Proposed\\(simultaneous learning)\end{tabular}} & VGG16 &  & 0.2795 & 0.2727 & 0.3256 & 0.3320 & 0.2731 \\
 & ResNet50 &  & 0.3148 & 0.3056 & 0.3665 & 0.3838 & 0.3120 \\
 & ConvNeXT-T &  & \underline{0.3295} & \textbf{0.3338} & \textbf{0.3899} & \textbf{0.4105} & \textbf{0.3331} \\
 & HLAGCN &  & \textbf{0.3318} & \underline{0.3194} & \underline{0.3854} & \underline{0.3918} & \underline{0.3265} \\ \midrule
\multirow{4}{*}{\begin{tabular}[c]{@{}c@{}}Conventional\\(rating distribution learning)\end{tabular}} & VGG16 & \multirow{8}{*}{SROCC ↑} & 0.2283 & 0.2031 & 0.2925 & 0.2736 & 0.2311 \\
 & ResNet50 &  & 0.2582 & 0.2523 & 0.3148 & 0.3175 & 0.2527 \\
 & ConvNeXT-T &  & 0.2849 & 0.2810 & 0.3419 & 0.3425 & 0.2764 \\
 & HLAGCN &  & 0.2829 & 0.2737 & 0.3402 & 0.3393 & 0.2745 \\ \cmidrule(r){1-2} \cmidrule(l){4-8} 
\multirow{4}{*}{\begin{tabular}[c]{@{}c@{}} Proposed\\(simultaneous learning)\end{tabular}} & VGG16 &  & 0.2717 & 0.2701 & 0.3118 & 0.2939 & 0.2762 \\
 & ResNet50 &  & 0.3029 & 0.2994 & 0.3451 & 0.3400 & 0.3081 \\
 & ConvNeXT-T &  & \underline{0.3164} & \textbf{0.3261} & \underline{0.3639} & \textbf{0.3601} & \textbf{0.3277} \\
 & HLAGCN &  & \textbf{0.3210} & \underline{0.3136} & \textbf{0.3647} & \underline{0.3519} & \underline{0.3241} \\ \midrule
\multirow{4}{*}{\begin{tabular}[c]{@{}c@{}}Conventional\\(rating distribution learning)\end{tabular}} & VGG16 & \multirow{8}{*}{MAE ↓} & 0.1551 & 0.1531 & 0.0537 & 0.0638 & 0.0345 \\
 & ResNet50 &  & \textbf{0.1513} & 0.1492 & 0.0533 & 0.0609 & 0.0343 \\
 & ConvNeXT-T &  & \underline{0.1544} & 0.1499 & 0.0526 & 0.0595 & 0.0343 \\
 & HLAGCN &  & 0.1631 & 0.1555 & 0.0536 & 0.0622 & 0.0351 \\ \cmidrule(r){1-2} \cmidrule(l){4-8} 
\multirow{4}{*}{\begin{tabular}[c]{@{}c@{}} Proposed\\(simultaneous learning)\end{tabular}} & VGG16 &  & 0.1565 & 0.1462 & 0.0533 & 0.0606 & 0.0340 \\
 & ResNet50 &  & 0.1608 & 0.1439 & 0.0526 & \textbf{0.0565} & 0.0336 \\
 & ConvNeXT-T &  & 0.1550 & \textbf{0.1431} & \textbf{0.0520} & \underline{0.0572} & \textbf{0.0333} \\
 & HLAGCN &  & \underline{0.1544} & \underline{0.1436} & \textbf{0.0520} & 0.0575 & \textbf{0.0333} \\ \midrule
\end{tabular}
\caption{Results of subjectivity regression. The best and the second best cases are marked with bold faces and underlines, respectively.}
\label{tab:subjectiveresults}
\end{table*}

\section{Predicting image aesthetics}
\label{sec:experiment}

Based on our aesthetic modeling framework, we propose a method to train neural networks for prediction of image aesthetics, called \emph{simultaneous learning}. We aim to obtain the rating distribution from a trained neural network, so that we can perform any of the binary classification of the mean rating, mean rating regression, subjectivity regression, and rating distribution prediction. For this, it is sufficient to make the neural network predict $\alpha$ and $\beta$ of the fitted beta distribution. Our preliminary experiment showed that this approach improves the performance of subjectivity regression compared to the conventional approach where the rating distribution is predicted, but the performance of the other tasks is lowered instead. 

To solve this problem, we design the neural network model so as to predict both the rating distribution and the shape parameters of the fitted beta distribution. For training of the model, we formulate three losses, $L_1$, $L_2$, and $L_3$ (Fig. \ref{fig:loss}). First, $L_1$ is calculated between the ground truth rating distribution ($R_T$) and the predicted distribution ($R_P$) in terms of EMD, i.e.,
\begin{equation}
L_{1}=\textrm{EMD}(R_P,R_T).
\label{eq:l1}
\end{equation}
Second, $L_2$ is the root mean squared log error (RMSLE) between the ground truth shape parameters (i.e., the shape parameters of the beta distribution fitted to the ground truth rating distribution) ($B_T$) and the predicted shape parameters ($B_P$):
\begin{equation}
L_{2}=\textrm{RMSLE}(B_P,B_T).
\label{eq:l2}
\end{equation}
Here, RMSLE is used instead of mean squared error (MSE) or root mean squared error (RMSE) in order to effectively handle the shape parameters that become larger by orders of magnitude occasionally for some images. Third, the predicted shape parameters $B_P$ are used to build a rating distribution, which is compared to the predicted rating distribution $R_P$:
\begin{equation}
L_{3}=\textrm{EMD}(R_P,B2R(B_P)),
\label{eq:l3}
\end{equation}
where $B2R(B_P)$ represents the process of converting $B_P$ to a rating distribution. This loss reduces the discrepancy between the two sets of model outputs, i.e., the predicted distribution and the beta distribution described by the predicted shape parameters. Finally, the total loss for our simultaneous learning is given by a weighted sum of the three losses, i.e.,
\begin{equation}
L=w_{1}\times L_{1}+w_{2}\times w_{b}\times L_{2}+w_{3}\times L_{3},
\label{eq:loss}
\end{equation}
where $w_1$, $w_2$, and $w_3$ are the weights for the three losses, respectively, with $w_1+w_2+w_3=1$, and $w_b$ is used to compensate for the different scales of EMD and RMSLE.

\section{Experiments}
\label{experiments}
\subsection{Dataset}
\label{sec:dataset}
We use the AVA dataset \cite{murray2012ava}. It contains photos and their aesthetic ratings from challenges of DPChallenge\footnote{http://www.dpchallenge.com}. The dataset is composed of 256,000 images. 236,000 images are for training and 20,000 are for test.

\subsection{Backbone models}
We use popular generic CNN models and a latest image aesthetic assessment model as backbone models for our experiments. The former includes VGG16 \cite{simonyan2015very}, ResNet-50 \cite{he2016deep}, and ConvNeXT-T \cite{liu2022convnet}, and the latter corresponds to the hierarchical layout-aware graph convolutional network (HLAGCN) \cite{she2021hierarchical}. HLAGCN models the complex relations among interesting regions in the input image using a graph convolutional network.

\subsection{Approaches}
In the proposed method using the simultaneous learning, a model produces 12 output values (ten for a rating distribution and two for the shape parameters of a beta distribution). For comparison, we employ the conventional approach that predicts only the rating distribution \cite{talebi2018nima, Chen_2020_CVPR, ching2020learning, she2021hierarchical}.

\begin{table*}[t]
\centering
\begin{tabular}{@{}cc|c|ccc|cc@{}}
\toprule
\multirow{2}{*}[-0.4em]{Approach} & \multirow{2}{*}[-0.4em]{Backbone} & Binary classification & \multicolumn{3}{c|}{Mean rating} & \multicolumn{2}{c}{Distribution}\\ \cmidrule(l){3-8} 
 & & Accuracy ↑& \multicolumn{1}{c}{PLCC ↑} & \multicolumn{1}{c}{SROCC ↑} & \multicolumn{1}{c|}{MAE ↓}  & \multicolumn{1}{c}{EMD ↓} & \multicolumn{1}{c}{KLD ↓} \\ \midrule
\multirow{4}{*}{\begin{tabular}[c]{@{}c@{}}Conventional\\(rating distribution learning)\end{tabular}} & VGG16 & 0.7818 & 0.6632 & 0.6582 & 0.447 & 0.0484 & 0.0936 \\
 & ResNet50 & 0.7936 & 0.697 & 0.6931 & 0.4282 & 0.0465 & 0.0885 \\
 & ConvNeXT-T & 0.7942 & \textbf{0.7225} & \textbf{0.7207} & 0.4196  & \textbf{0.0456} & \underline{0.0841}\\
 & HLAGCN & \textbf{0.7965} & 0.7031 & 0.7104 & 0.4163 & 0.0459 & \textbf{0.0825} \\ \midrule
\multirow{4}{*}{\begin{tabular}[c]{@{}c@{}}Proposed\\(simultaneous learning)\end{tabular}} & VGG16 & 0.7757 & 0.6513 & 0.6443 & 0.4536 & 0.0497 & 0.1044 \\
 & ResNet50 & 0.7933 & 0.6931 & 0.6905 & 0.425 & 0.0468 & 0.0963 \\
 & ConvNeXT-T & \underline{0.7957} & \underline{0.7156} & \underline{0.7110} & \textbf{0.4151} & 0.0462 & 0.0891 \\
 & HLAGCN & 0.7942 & 0.7077 & 0.7054 & \underline{0.4158}  & \underline{0.0458} & 0.0876\\ \bottomrule
\end{tabular}
\caption{Results of binary classification, mean rating regression, and rating distribution prediction. The best and the second best cases are marked with bold faces and underlines, respectively.}
\label{tab:non-subjectiveresults}
\end{table*}

\subsection{Performance measures}

We consider various tasks of image aesthetic assessment, and use appropriate performance measures for each task by following the previous studies \cite{talebi2018nima, kang2019predicting}.

Our simultaneous learning method produces both a predicted rating distribution and a predicted beta distribution (in the form of its shape parameters). Thus, both can be used to measure the performance of the tasks. In our experiments, the predicted rating distribution is used for binary classification, mean rating regression, and rating distribution prediction. For subjectivity regression, the predicted subjectivity metrics are computed from the predicted beta distribution, which yields better performance. The predicted AesU is obtained directly from the predicted shape parameters using Eq. (\ref{eq:bdu2}). In the case of the conventional approach predicting only the rating distribution, the existing subjectivity metrics are computed using the predicted rating distribution and AesU is obtained after fitting to a beta distribution.

For the binary classification task, the classification accuracy is used. The threshold of the two classes is set to a score of 5 in the scale of 1 to 10 as in \cite{lu2014rapid, Mai_2016_CVPR, Liu_2019_CVPR_Workshops, sheng2020revisiting}. The performance of mean rating regression is measured in terms of mean absolute error (MAE), Pearson linear correlation coefficient (PLCC), and Spearman rank-order correlation coefficient (SROCC) between the ground truth mean ratings and the predicted mean ratings. The same measures are also used for the subjectivity regression task. For distribution prediction, we use EMD and the Kullback-Leibler divergence (KLD) between the ground truth and predicted distributions. 

\subsection{Implementation details}
All experiments are conducted on a PC that has AMD Ryzen 5 5600x CPU, 128GB of RAM, NVIDIA Geforce RTX 3090 24GB GPU, and Microsoft Windows 10. We use Python 3.9.7, PyTorch 1.10.1, CUDA 11.3, and cuDNN 8.0.

We divide the training dataset further into 223,000 training images and 13,000 validation images. The images are resized to $256 \times 256$ ($336 \times 336$ for HLAGCN) and randomly cropped to $224 \times 224$ ($300 \times 300$ for HLAGCN). and a random horizontal flip is applied for data augmentations.

In the simultaneous learning, we set $w_{1}=0.4$, $w_{2}=0.5$, and $w_{3}=0.1$, indicating the relative importance of the three losses, and $w_{b}=0.2$.

To train the models, we use the SGD optimization with a batch size of 48, a Nesterov momentum paramenter of 0.9 and a weight decay parameter of $5\times10^{-4}$. The learning rate is initially set to $5\times10^{-3}$ and reduced by 5\% every 10 epochs. The models are trained for 100 epochs, but if the validation loss does not decrease for 10 epochs, the learning is stopped.

\subsection{Results}

Table \ref{tab:subjectiveresults} shows the results of the subjectivity regression task for STD, MAD, MED, DUD, and AesU. The proposed simultaneous learning approach shows improved performance compared to the conventional approach in all cases except that the proposed approach yields the second best performance in the case of STD with MAE. When the backbone models are compared, ConvNeXT-T performs the best, and HLAGCN also performs similarly well. For ConvNeXT-T, the improvement by the proposed approach over the conventional one is about 5\% on average, with the maximum improvement by about 21\% for AesU in terms of PLCC. These results demonstrate that the proposed simultaneous learning approach using the three losses is effective to learn the subjectivity of image aesthetics. In particular, the proposed subjectivity modeling plays a key role by reducing the noise in the raw rating distribution.

Table \ref{tab:non-subjectiveresults} summarizes the results of the other tasks, i.e., binary classification, mean rating regression, and rating distribution prediction. The proposed approach shows slightly lowered performance than the conventional approach but only marginally. The performance decrease by the proposed approach for ConvNeXT-T is only 1.4\% on average. Note that this performance decrease is because we tune our approach to maximize the performance of the subjectivity regression task.

\begin{table*}[t]
\centering

\begin{tabular}{@{}c|c|cc|c|cc|cc|cc|cc|cc@{}}
\toprule
\multirow{3}{*}[-0.65em]{Approach} & \multirow{2}{*}[-0.25em]{\begin{tabular}[c]{@{}c@{}}Binary\\ classif.\end{tabular}} & \multicolumn{2}{c|}{\multirow{2}{*}[-0.4em]{Mean rating}} & \multirow{2}{*}[-0.4em]{Dist.} & \multicolumn{10}{c}{Subjectivity} \\ \cmidrule(l){6-15} 
 &  & \multicolumn{2}{c|}{} &  & \multicolumn{2}{c|}{STD} & \multicolumn{2}{c|}{MAD} & \multicolumn{2}{c|}{MED} & \multicolumn{2}{c|}{DUD} & \multicolumn{2}{c}{AesU} \\ \cmidrule(l){2-15} 
 & Accuracy & LCC & MAE & EMD & LCC & MAE & LCC & MAE & LCC & MAE & LCC & MAE & LCC & MAE \\ \midrule
Rating  & \underline{0.794} & \textbf{0.723} & \underline{0.419}& \textbf{0.046} & 0.293 & \underline{0.154} & 0.290 & 0.150 & 0.364 & 0.053 & \underline{0.390} & \underline{0.060} & 0.275 & 0.034 \\
Beta & 0.782 & 0.682 & 0.439& 0.050 & \textbf{0.339} & \textbf{0.152} & \underline{0.332} & \underline{0.144} & \underline{0.383} & \textbf{0.052} & 0.377 & 0.060 & \textbf{0.336} & \textbf{0.033} \\
Both & \textbf{0.796} & \underline{0.716} & \textbf{0.415} & \textbf{0.046} & \underline{0.330} & 0.155 & \textbf{0.334} & \textbf{0.143} & \textbf{0.390} & \textbf{0.052} & \textbf{0.411} & \textbf{0.057} & \underline{0.333} & \textbf{0.033} \\ \bottomrule
\end{tabular}
\caption{Comparison of the cases where the neural network models output either a predicted rating distribution or a predicted beta distribution (in the form of shape parameters), or both (proposed simultaneous learning). The best and the second best cases are marked with bold faces and underlines, respectively.}
\label{tab:loss1}
\end{table*}

\begin{table*}[t]
\centering

\begin{tabular}{@{}c|c|cc|c|cc|cc|cc|cc|cc@{}}
\toprule
\multirow{3}{*}[-0.65em]{Fitting} & \multirow{2}{*}[-0.25em]{\begin{tabular}[c]{@{}c@{}}Binary\\ classif.\end{tabular}} & \multicolumn{2}{c|}{\multirow{2}{*}[-0.4em]{Mean rating}} & \multirow{2}{*}[-0.4em]{Dist.} & \multicolumn{10}{c}{Subjectivity} \\ \cmidrule(l){6-15} 
 &  & \multicolumn{2}{c|}{} &  & \multicolumn{2}{c|}{STD} & \multicolumn{2}{c|}{MAD} & \multicolumn{2}{c|}{MED} & \multicolumn{2}{c|}{DUD} & \multicolumn{2}{c}{AesU} \\ \cmidrule(l){2-15} 
 & Accuracy & LCC & MAE & EMD & LCC & MAE & LCC & MAE & LCC & MAE & LCC & MAE & LCC & MAE \\ \midrule
w/ & 0.796 & \textbf{0.716} & \textbf{0.415} & \textbf{0.046} & \textbf{0.330} & \textbf{0.155} & \textbf{0.334} & \textbf{0.143} & \textbf{0.390} & \textbf{0.052} & \textbf{0.411} & 0.057 & \textbf{0.333} & \textbf{0.033} \\
w/o & \textbf{0.797} & 0.712 & \textbf{0.415} & \textbf{0.046} & 0.283 & 0.163 & 0.277 & 0.145 & 0.352 & 0.053 & 0.403 & \textbf{0.056} & 0.288 & 0.034 \\ \bottomrule
\end{tabular}
\caption{Comparison of the two ways measuring $L_2$ (without or with fitting to beta distributions) in our simultaneous learning method. The better case is highlighted.}
\label{tab:loss2}
\end{table*}

\subsection{Ablation study}
We further conduct additional experiments with the best-performing model (ConvNeXT-T) to examine the effectiveness of the simultaneous learning and the way to compute the loss of the predicted beta distribution.

\begin{figure}[t]
\centering
\includegraphics[width=\columnwidth]{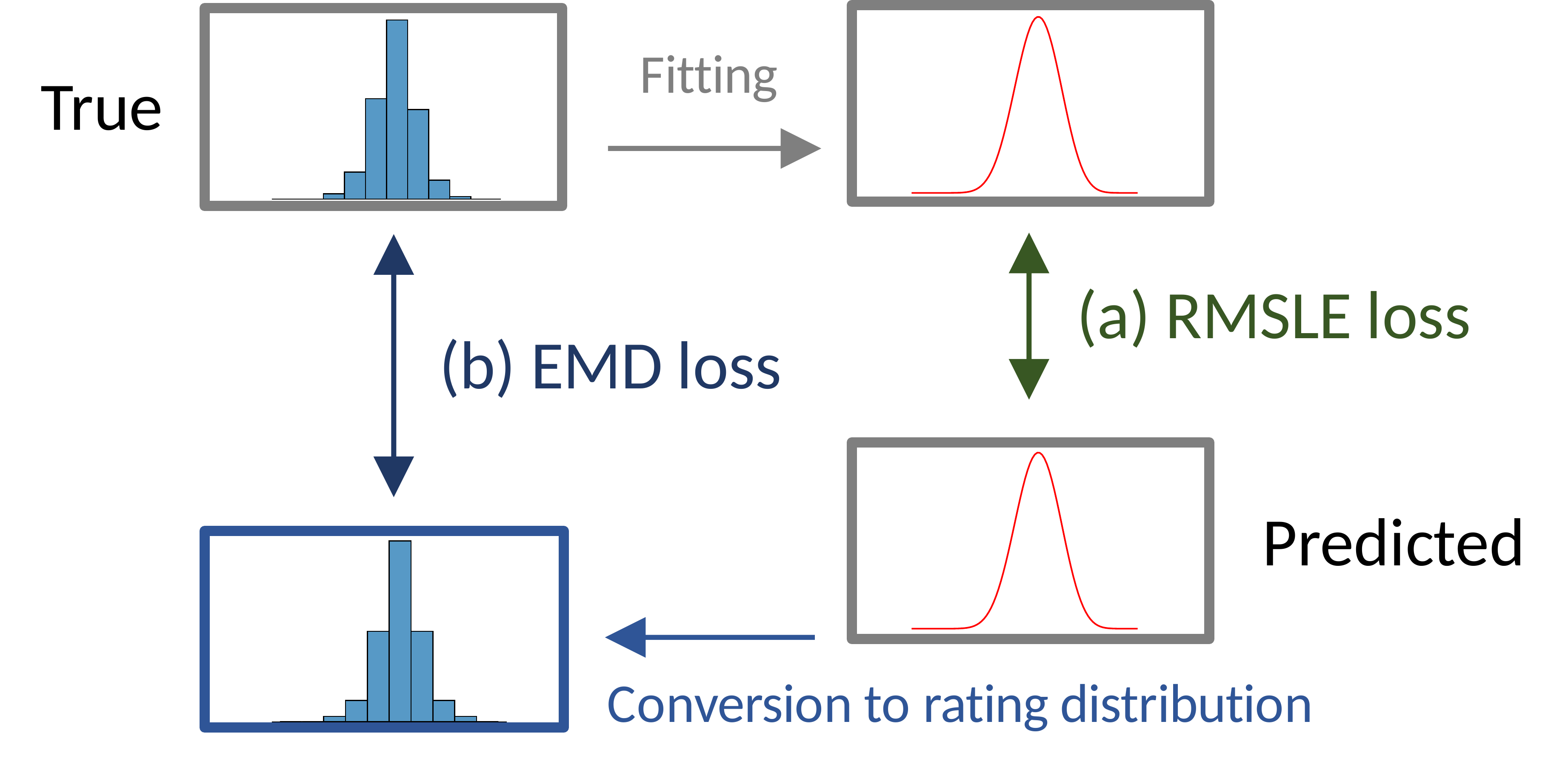}  
\caption{Two ways of measuring the loss of the predicted beta distribution (i.e., $L_2$). (a) As in Fig. \ref{fig:loss}, RMSLE is measured between the predicted shape parameters and the ground truth shape parameters (obtained by fitting to the ground truth rating distribution). (b) The predicted shape parameters are converted to a rating distribution, which is compared to the ground truth rating distribution in terms of EMD.}
\label{fig:betaloss}
\end{figure}

\subsubsection{Impact of simultaneous learning.}

In our simultaneous learning method, the neural network models predict both the rating distribution and the shape parameters of the beta distribution. In the conventional approach, only the rating distribution is predicted. We additionally test the case where the models predicts only the shape parameters of the beta distribution. The results are shown in Table \ref{tab:loss1}. Predicting only the beta distribution loses details of the original rating distribution, resulting in a lowered EMD in distribution prediction. However, the case of predicting the beta distribution outperforms the case of predicting the rating distribution in subjectivity regression, indicating the usefulness of our subjectivity modeling using beta distributions. Overall, the proposed simultaneous learning shows the best performance among the three cases.

\subsubsection{Impact of fitting to beta distribution.}

In our loss function, $L_2$ is measured between the beta distribution fitted to the ground truth rating distribution and the predicted beta distribution (using their shape parameters). Instead, it can be also measured without fitting, i.e., between the ground truth rating distribution and the rating distribution converted from the predicted beta distribution (Fig. \ref{fig:betaloss}). In this case, the weights ($w_1$, $w_2$, and $w_3$) are kept the same, but $w_b$ is set to 1 because any scale difference between the losses do not exist. The two cases are compared in Table \ref{tab:loss2}. The performance of subjectivity regression is degraded significantly without fitting, while the performance of the other tasks remains almost the same.

\subsection{Application to aesthetic image recommendation}
\label{sec:application}

\begin{figure}[t]
\centering
\includegraphics[width=0.6\columnwidth]{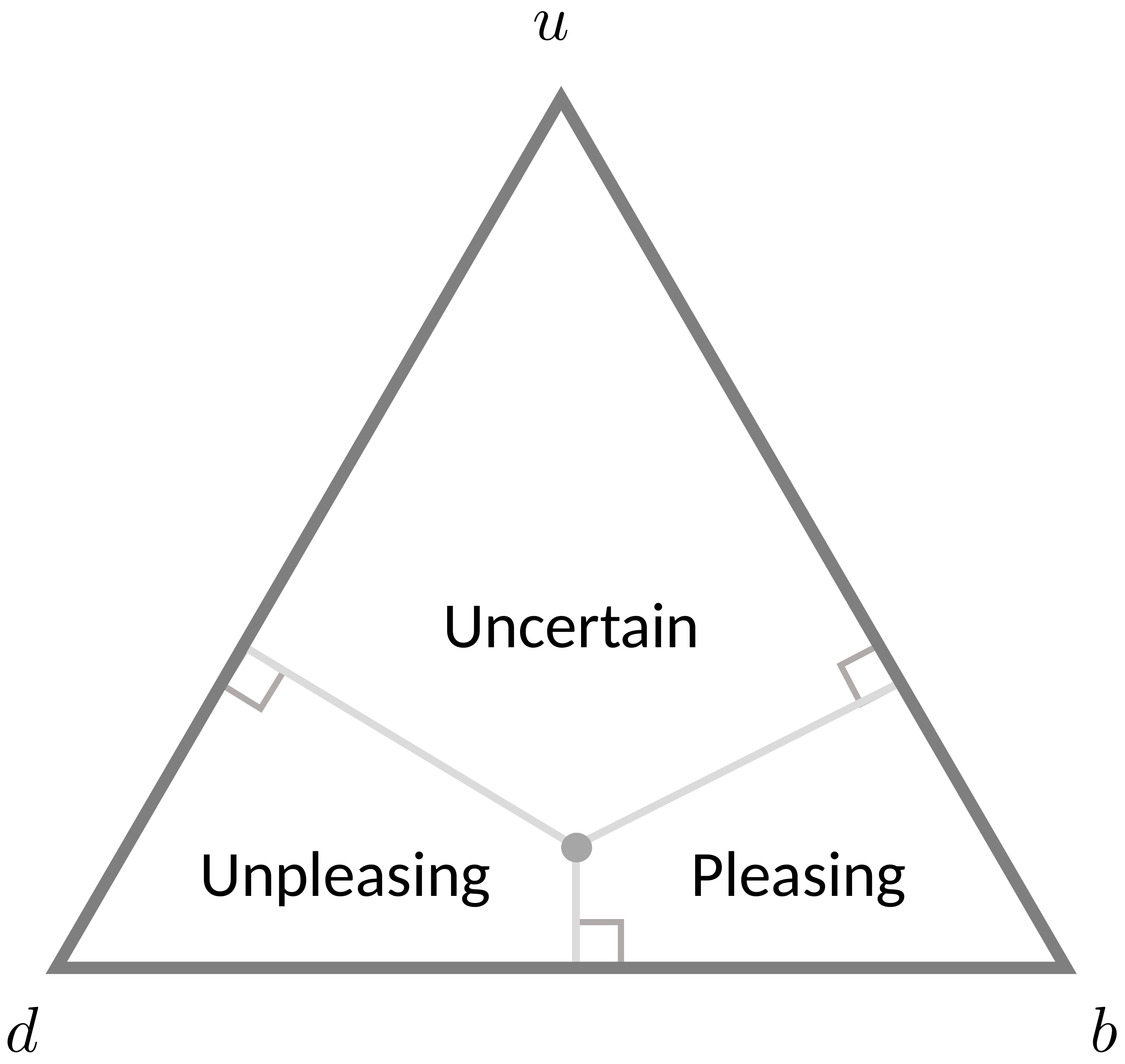}  
\caption{Class boundaries of the ternary classification on the equilateral triangle of $b$, $d$, and $u$. The center point is set to the median value of each component for the training dataset, which is $(b,d,u)=(0.419, 0.444, 0.137)$.}
\label{fig:ternary}
\end{figure}

Finally, we illustrate an application scenario where our subjectivity modeling can be applied.

Consider an image recommender application, where the image aesthetics of the recommended images is considered important. Even if an image is predicted to be aesthetically pleasing in terms of the mean rating, if a nonnegligible proportion of users would disagree on the prediction, then it would be better not to recommend the image to users in order to maximize the users' satisfaction. 

For this, the proposed framework can be used to filter out the images that are predicted to have high subjectivity levels. We perform ternary classification to classify an image as pleasing, unpleasing, or uncertain, based on the three probabilities obtained from our subjectivity modeling framework. The boundaries between the three classes are defined on the equilateral triangle of the three probabilities as shown in Fig. \ref{fig:ternary}. The center point where the boundaries meet is set by the median values of $b$, $d$, and $u$ for the training dataset.\footnote{The center point can be simply set to $(b,d,u)=(1/3, 1/3, 1/3)$, but the number of images in each class becomes significantly unbalanced. Thus, we use the median values.} Then, the recommendation rule is to recommend only the images classified to be pleasing. For comparison, we also consider the baseline rule to recommend an image if it is predicted to be aesthetically pleasing from binary classification.

We simulate the two rules on the test data of the AVA dataset using the trained ConvNeXT-T model. Their performance is measured by the satisfaction ratio, which is calculated as the average proportion of raters who voted scores higher than the threshold of binary classification over the recommended images. The obtained satisfaction ratios are 58.26\% and 61.79\% for the baseline rule and our rule, respectively. This improvement demonstrates the effectiveness of our framework in this application.

\section{Conclusion}
\label{sec:conclusion}

We proposed a unified probabilistic framework for modeling image aesthetics, particularly considering proper quantification of the subjectivity. The framework allowed us to model the rating distribution as a beta distribution and to obtain the probabilities of being definitely pleasing, being definitely unpleasing, and being uncertain simultaneously. The probability of being uncertain was used to define AesU, which is an intuitive subjectivity metric. We also proposed the simultaneous learning method to improve the performance of subjectivity prediction, which was confirmed by the experiments. Finally, we showed that our framework can be effective for aesthetics-based image recommendation.

\printbibliography

\end{document}